\documentclass[11pt,a4paper]{article}
\usepackage{times,latexsym}
\usepackage{url}
\usepackage[T1]{fontenc}

\usepackage{multirow}
\usepackage{xcolor}
\usepackage{amsmath}
\usepackage{pifont}
\usepackage{graphicx}
\usepackage{lipsum}
\usepackage{newtxmath}
\usepackage[normalem]{ulem}
\graphicspath{ {./img/} }

\usepackage{rparxiv}
\usepackage{gb4e}
\noautomath

\newlength{\rindent}
\usepackage{xspace,mfirstuc,tabulary}

\newif\iftaclinstructions
\taclinstructionsfalse % AUTHORS: do NOT set this to true
\iftaclinstructions

\renewcommand{\anonsubtext}{(No author info supplied here, for consistency with
TACL-submission anonymization requirements)}
\newcommand{\instr}
\fi

\iftaclpubformat % this "if" is set by the choice of options

\else

\fi

\definecolor{sentencegood}{RGB}{0, 128, 0}
\definecolor{sentencebad}{RGB}{140, 0, 0}
\newcommand{\textgood}[1]{\textcolor{sentencegood}{#1}}
\newcommand{\textbad}[1]{\textcolor{sentencebad}{#1}}
\newcommand{\textbg}[1]{\textcolor{blue!60!black}{#1}}
\newcommand{\textsent}[1]{\textcolor{orange!80!black}{#1}}
\newcommand{\revisit}[1]{\textcolor{red}{#1}}

\title{Prompts and probabilities in the linguistic assessment of language models: Evidence from semantic focus-sensitivity%\\ \Taclpaper
}

\author{
  Robert Pasternak
  \\
  Leibniz-Centre General Linguistics (ZAS)
  \\
  \texttt{mail@robertpasternak.com}
}

\date{}

\begin{document}
\maketitle
\begin{abstract}
  As language models (LMs) have improved, an increasingly common means of evaluating their linguistic competence is through explicit prompting (e.g., ``Which sentence is more acceptable?''). Prior work indicates that such prompting underperforms compared to standard token-probability-based methods in smaller models, but improves in larger models. We test this in a new domain: \emph{semantic focus-sensitivity}, responsible for contrasts like \emph{Lucy only SWAM yesterday} vs.\ \emph{Lucy only swam YESTERDAY}. Testing models of various sizes using text-based focus markers like boldface and italics, we find universal above-chance performance for token probabilities. Prompting results mostly align with previous findings, but in some larger models they outperform probabilities by a wide margin. We propose that this is because LMs are not prone to produce out-of-the-blue focus marking, which muddies results from token probabilities in a way that does not affect prompting. We test this by replacing standard focus markers with randomly generated ones; token probabilities still beat chance but show significantly greater performance decrease compared to prompting. These results suggest that metalinguistic prompting, while non-trivial for LMs, can also be more robust when testing stylistically atypical data. We argue that this has important consequences for the assessment of LMs' language competence, even if LMs do not perform metalinguistic judgments through human-like introspection.\footnote{Code and data are available at: \iftaclpubformat{\url{https://github.com/robpasternak/prompts-probs-focus}}\else{\revisit{[anonymized; GitHub URL will be provided]}}\fi.}

\end{abstract}

\section{Introduction}

The linguistic competence of language models (LMs) is often assessed by means of their output token probabilities: roughly, well-formed sentences are \emph{ceteris paribus} expected to have higher token or string probabilities than ill-formed ones. While some prior work has used token probabilities to directly calculate acceptability scores on a per-sentence basis (\citealt{ClarkEtAl2013,LauEtAl2015}), a more common technique is to compare the probabilities assigned to \textbf{minimal pairs}, near-identical pairs of sentences that differ in a single, targeted way, such as \emph{Mary is/*are sad} (e.g., \citealt{LinzenEtAl2016,GulordavaEtAl2018,MarvinLinzen2018,FutrellEtAl2019,WarstadtEtAl2020}). Comparison of minimal pairs is common in theoretical and experimental research on human language (see e.g.\ \citealt{SprouseEtAl2013}), and is useful in NLP research because it helps control for irrelevant factors that affect token or string probabilities, such as sentence length and the general frequency of individual tokens.

Thanks to major improvements in LM capabilities, an increasingly common alternative means of evaluating LMs' linguistic competence is to simply present the relevant sentence(s) as part of a prompt to the model, along with a request for a judgment pertaining to well-formedness (\citealt{DentellaEtAl2023,Katzir2023,Mahowald2023,MurphyEtAl2025}) or semantic/pragmatic truth or acceptability (e.g.\ \citealt{KamathEtAl2024,TsvilodubEtAl2024}). Such metalinguistic prompting is similar to judgment tasks commonly used in experimental research on humans, making it easy to construct studies comparing human and LM responses to more or less the same input.

However, research directly comparing metalinguistic prompting to token probabilities shows very different performance between the two methods. Weaker LMs generally perform much worse when assessed by prompting than they do when assessed through string probabilities (\citealt{HuLevy2023,HuEtAl2024,KaufEtAl2024}), leading to significant underestimation of these models' language competence. \citet{HuFrank2024} observe that this gap in accuracy shrinks as models get larger and more capable; they attribute this to the auxiliary task demands of metalinguistic judgments, which stronger models are better able to cope with. \citet{SongEtAl2025}, meanwhile, find that with sufficiently large models, metalinguistic judgments can slightly but significantly outperform string probabilities.

In this paper we report two studies extending this foundational methodological research to a new empirical domain. Our results suggest a novel generalization about the accuracy of prompting versus string probabilities: while metalinguistic prompting is indeed only accessible to sufficiently large models, it can also be more robust than string probabilities when evaluating sentences that deviate from a model's stylistically typical outputs.

Our empirical domain is \textbf{semantic focus-sensitivity}, a linguistic phenomenon in which the truth or pragmatic acceptability of a sentence depends on the placement of stress. By way of illustration, consider an adverb like \emph{only}. If Mary gave one book to her teacher and one to her friend (and nothing to anyone else), then (\ref{ex:only-true}) is true and (\ref{ex:only-false}) false, despite differing only in focus placement (\citealt{Jackendoff1972,Rooth1985}):
\begin{exe}
  \ex   Mary only gave her teacher a BOOK.\label{ex:only-true}
  \ex   Mary only gave her TEACHER a book.\label{ex:only-false}
\end{exe}
As discussed in greater detail in Section \ref{sec:background}, semantic focus-sensitivity arises in a variety of environments in human language. Notably, in many cases the appropriateness of focus placement is dependent on context: \eqref{ex:only-true} and \eqref{ex:only-false} are both fine sentences in and of themselves, and it is only the context that makes \eqref{ex:only-true} acceptable and \eqref{ex:only-false} not.

Our goals here are twofold. First, we wish to determine whether LMs have learned focus-sensitivity. And second, we compare and contrast string probabilities and prompting as methods of linguistically evaluating LMs. We test instruction-tuned models of various sizes from three model families---Gemma 4 (\citealt{Gemma4}), Qwen3 (\citealt{Qwen3}), and Qwen3.5 (\citealt{Qwen3.5})---on a manually created 282-item minimal pair dataset featuring various kinds of semantic focus-sensitivity, comparing the results using string probabilities to those using two types of prompts: \textbf{forced choice}, in which the LM is asked to choose which of the two sentences in a pair is more acceptable; and \textbf{yes/no}, in which the LM is presented with one sentence at a time and asked for a yes-or-no acceptability judgment. These methods are illustrated in Table \ref{fig:methods-illustration}.

\begin{table*}
  \footnotesize
  \begin{center}
  \begin{tabular}{r | p{.15\textwidth} | p{.3\textwidth} | p{.3\textwidth}}
    & \textbf{String Prob.} & \textbf{Forced Choice} & \textbf{Yes/No}\\
    \hline

    \textbf{\textsc{system}} & \multicolumn{3}{ c }{You are a helpful assistant.}\\
    \hline
    
    \textbf{\textsc{user}} &
    \textbg{Who did Phil give a book to?} &
    You will be presented with a context followed by two sentences, A and B. Your job is to respond with which sentence is better in the context provided.\vspace{\baselineskip}

    CONTEXT:

    \textbg{Who did Phil give a book to?}\vspace{\baselineskip}

    SENTENCE A: \textsent{Phil gave a book to the SENATOR.}

    SENTENCE B: \textsent{Phil gave a BOOK to the senator.}&
    You will be presented with a context followed by a sentence. Your job is to respond with "Yes" or "No", indicating whether the sentence is acceptable in the context provided.\vspace{\baselineskip}

    CONTEXT:

    \textbg{Who did Phil give a book to?}\vspace{\baselineskip}

    SENTENCE: \textsent{Phil gave a book to the SENATOR.}\\
    \hline
    \textbf{\textsc{assistant}}&
    \textgood{Phil gave a book to the SENATOR.} / \textbad{Phil gave a BOOK to the senator.} &
    My answer is \textgood{A} / \textbad{B} &
    My answer is \textgood{Yes} / \textbad{No}\\
  \end{tabular}
\end{center}
  \normalsize
  \caption{Illustration of string probability, forced choice, and yes/no evaluation methods.\label{fig:methods-illustration}}
\end{table*}

In Experiment 1, discussed in Section \ref{sec:exp-1}, we test how models perform using eight relatively standard methods for text-based focus-marking: all caps; HTML and Markdown boldface, italics, and emphasis; and HTML underlining. We find that for all models, string probabilities perform significantly above chance, suggesting substantial LM acquisition of semantic focus-sensitivity.

Regarding the methodological comparison, in certain respects the results accord nicely with prior work: prompting methods perform poorly for small models and gradually improve as models get bigger, with forced choice outperforming yes/no prompting (cf.\ \citealt{PanicksseryEtAl2024}). However, in the case of the larger Gemma 4 models forced choice prompting not only beats string probabilities, but does so by a wide margin. We hypothesize that this is because the models are not prone to produce out-of-the-blue focus marking, which lowers the general probability of unsolicited focus marking and muddies string probability results. But for models that are strong enough to perform forced choice, such stylistic unnaturalness is less of a hindrance, since (\textsc{i}) the focus marking is in the input, rather than the output; and (\textsc{ii}) the side-by-side comparison of paired sentences clarifies the criteria of evaluation, hence why forced choice also outperforms yes/no prompting.

To test this, in Experiment 2 (Section \ref{sec:exp-2}) we replace standard focus markers with randomly generated focus bracketing (e.g., \emph{Mary only gave her teacher a \$:\#book\#:\$}), on the assumption that these deviate even further from the models' expected outputs. While we unsurprisingly find general accuracy decreases, string probabilities continue to beat chance for all models, suggesting that what the models have learned is more general and not confined to a closed class of canonical text-based focus markers. However, string probabilities also show a proportionally greater decline in accuracy than forced choice prompting, with the number of instances of forced choice significantly outperforming string probabilities more than doubling. This supports the hypothesis that while forced choice prompting is more difficult for a model to acquire, it can also be more robust than string probabilities when evaluating sentences that deviate from a model's stylistic inclinations.

In Section \ref{sec:discussion} we discuss the consequences of these results for the linguistic assessment of LMs. This is especially pertinent because recent work suggests that the language knowledge LMs use to make metalinguistic judgments may be distinct from the implicit knowledge they use to generate outputs, inviting questions about which is the true measure of an LM's linguistic competence, regardless of comparative accuracy (\citealt{SongEtAl2025}). However, we argue that even if we accept these results at face value, there needn't be one ``true'' indicator of an LM's language abilities: it is equally plausible that these methods test distinct but coequal facets of the core language modeling task, loosely analogous to the contrast between production (string probabilities) and comprehension (prompting). From this perspective, prompting and probabilities complement each other, and studying both promises to give a more holistic picture of LM competence than can be achieved through one methodology alone.

\section{Focus-sensitive environments\label{sec:background}}

As was mentioned in the introduction, focus-sensitivity is common across a wide range of environments. We have already seen one example: association with \emph{only}, as exemplified by the contrast between \eqref{ex:only-true} and \eqref{ex:only-false}. Another example is \emph{question-answer congruence}, which ties the use of focus in an answer to its semantic relationship with the question (\citealt{vonStechow1991,Rooth1992}). Thus, in response to \emph{Who did Phil give a book to?}, (\ref{ex:qa-good}) is fine, while (\ref{ex:qa-bad}) is odd:
\begin{exe}
  \ex   Phil gave a book to the SENATOR.\label{ex:qa-good}
  \ex   Phil gave a BOOK to the senator.\label{ex:qa-bad}
\end{exe}

Yet another example is the focus-sensitivity of additive \emph{too}: as a continuation of \emph{The doctor greeted the nurse}, (\ref{ex:too-good}) is acceptable, while (\ref{ex:too-bad}) is not (\citealt{Kadmon2001,Kripke2009}):
\begin{exe}
  \ex   and the PATIENT greeted the nurse, too.\label{ex:too-good}
  \ex   and the patient greeted the NURSE, too.\label{ex:too-bad}
\end{exe}

A fourth example of focus-sensitivity can be seen in what \citet{Lewis1975} calls \emph{adverbs of quantification}, such as \emph{always} and \emph{usually} (\citealt{Rooth1985}, \citealt{Partee1991}, \citealt{Krifka1992}, \citealt{vonFintel1994}). If Sarah rarely runs, but when she does she opts for a morning run, then \eqref{ex:adv-qt-good} is true, while \eqref{ex:adv-qt-bad} is not:
\begin{exe}
  \ex   Sarah usually runs in the MORNING.\label{ex:adv-qt-good}
  \ex   Sarah usually RUNS in the morning.\label{ex:adv-qt-bad}
\end{exe}

Finally there is contrastive focus, such as so-called \emph{contrastive negation}: if someone claims that Sarah bought a green shirt when she actually bought a red one, then \eqref{ex:contr1-good} is a good response, while \eqref{ex:contr1-bad} is not (\citealt{Jackendoff1972}):
\begin{exe}
  \ex   Sarah didn't buy the GREEN shirt.\label{ex:contr1-good}
  \ex   Sarah didn't buy the green SHIRT.\label{ex:contr1-bad}
\end{exe}
Contrastive focus often comes in pairs. Thus, \eqref{ex:contr1-good} is better followed by \eqref{ex:contr2-good} than \eqref{ex:contr2-bad}:
\begin{exe}
  \ex   {\dots}she bought a RED shirt.\label{ex:contr2-good}
  \ex   {\dots}she bought a red SHIRT.\label{ex:contr2-bad}
\end{exe}

These do not constitute the full range of focus-sensitive environments (see \citealt{Kadmon2001,BeaverClark2008} for more). However, we consider the environments discussed above to be both diverse and commonplace enough to constitute a reasonable test domain for LM competence. With this in mind, we now turn to our experiments.

\section{Experiment 1: Standard focus markers\label{sec:exp-1}}

\subsection{Methodology}

We tested LMs' semantic focus-sensitivity using a manually created dataset consisting of 282 minimal pairs, representing each kind of focus-sensitivity discussed above: association with \emph{only} (\textsc{only}, 52 items), question-answer congruence (\textsc{qa}, 50 items), association with \emph{too} (\textsc{too}, 50 items), adverbs of quantification (\textsc{adv\_qt}, 40 items), contrastive negation (\textsc{contr1}, 40 items), and contrastive focus following negation as in \eqref{ex:contr2-good} (\textsc{contr2}, 50 items). In order to be able to test a variety of different text-based focus markers, the dataset was created as a ``seed'' with focus indicated by bracketing; bracketed material could then be replaced with any of a variety of focus markers.

Every item came with either a \textbf{background}, i.e., information that is required to assess the acceptability of a sentence in the minimal pair; or a \textbf{prefix}, the beginning of a statement for which the minimal pair sentences would serve as continuations. \textsc{only} and \textsc{adv\_qt} items came with informational backgrounds that entailed the good sentence, but not the bad sentence.\footnote{More specifically, for these items the backgrounds were \emph{neutral} with respect to the truth of the bad sentence. For some models we also tried an alternate version of the dataset where the background explicitly \emph{contradicted} the bad sentence; this led to worse results, which we attribute to the additional complexity required of the backgrounds.} \textsc{qa} items' backgrounds were simply the question to be answered, while \textsc{contr1} items had backgrounds that made only the good sentence pragmatically felicitous. \textsc{too} and \textsc{contr2} items had prefixes. Importantly, none of the backgrounds or prefixes had any focus marking themselves, so in all items focus marking was ``out-of-the-blue''. Examples of the six categories of item can be seen in Table \ref{tab:sample-items}.

\begin{table*}[t]
  \small
  \begin{center}
  \begin{tabular}{c | p{7.1cm} | p{6.1cm} }
    \textbf{Category} & \multicolumn{1}{c|}{\textbf{Background/Prefix}} & \multicolumn{1}{c}{\textbf{Sentences}}\\
    \hline
    \textsc{only} & \textbf{\textit{Background:}} Larry's Pub serves drinks at the bar. They do not serve drinks at other places. & \textgood{Larry's Pub only serves drinks at the BAR.}\newline\textbad{Larry's Pub only serves DRINKS at the bar.}\\
    \hline
    \textsc{qa} & \textbf{\textit{Background:}} Who did Phil give a book to? & \textgood{Phil gave a book to the SENATOR.}\newline\textbad{Phil gave a BOOK to the senator.}\\
    \hline
    \textsc{too} & \textbf{\textit{Prefix:}} A bonus was given to the senior engineer, & \textgood{and the JUNIOR engineer got a bonus, too.}\newline\textbad{and the junior engineer got a BONUS, too.}\\
    \hline
    \textsc{adv\_qt} & \textbf{\textit{Background:}} When Dolly buys rugs, it's with cash most of the time. & \textgood{Dolly mostly uses CASH to buy rugs.}\newline\textbad{Dolly mostly uses cash to buy RUGS.}\\
    \hline
    \textsc{contr1} & \textbf{\textit{Background:}} Sarah was considering buying a green shirt. She ended up buying a red one instead. & \textgood{Sarah didn't buy the GREEN shirt.}\newline\textbad{Sarah didn't buy the green SHIRT.}\\
    \hline
    \textsc{contr2} & \textbf{\textit{Prefix:}} Sandy didn't unlock the door. & \textgood{The door is still LOCKED.}\newline\textbad{The DOOR is still locked.}
  \end{tabular}
  \end{center}
  \normalsize
  \caption{Examples of the six categories of item in the dataset. For these examples, focus is marked with all caps.\label{tab:sample-items}}
\end{table*}

For Experiment 1, we tested eight anecdotally common ways of textually marking focus: all caps (standard in theoretical linguistics); HTML underlining; and HTML and Markdown boldface, italicized, and emphasized fonts. Thus, the full dataset consisted of 8 $\times$ 282 $=$ 2,256 pairs.

% \begin{table}
%   \begin{center}
%     \begin{tabular}{c | c}
%       \textbf{Focus Marker} & \textbf{Example}\\
%       \hline
%       \textsc{all\_caps} & the DOG\\
%       \textsc{html\_bold} & the <b>dog</b>\\
%       \textsc{html\_emph} & the <em>dog</em>\\
%       \textsc{html\_ital} & the <i>dog</i>\\
%       \textsc{html\_uline} & the <u>dog</u>\\
%       \textsc{mdown\_bold} & the **dog**\\
%       \textsc{mdown\_emph} & the *dog*\\
%       \textsc{mdown\_ital} & the \_dog\_\\
%     \end{tabular}
%   \end{center}
%   \caption{Focus markers tested in Experiment 1.\label{tab:focus-marking-exp-1}}
% \end{table}

We tested seventeen instruction-tuned models from three model families: Gemma 4 (\citealt{Gemma4}), Qwen3 (\citealt{Qwen3}), and Qwen3.5 (\citealt{Qwen3.5}). These model families were selected because each has a relatively wide range of model sizes. The tested models ranged in size from 600 million to 35 billion parameters. The Gemma 4 models tested were E2B, E4B, 12B, 26B-A4B, and 31B; the Qwen3 models were 0.6B, 1.7B, 4B, 8B, 14B, and 32B; and the Qwen3.5 models were 0.8B, 2B, 4B, 9B, 27B, and 35B-A3B. Notably, for Gemma 4 and Qwen3.5, some of the models are architecturally distinct enough from their peers to preclude a true apples-to-apples size comparison, whether that be due to a mixture-of-experts architecture (Gemma 4 26B-A4B, Qwen3.5 35B-A3B) or per-layer embeddings (Gemma 4 E2B, E4B). Thus, discussion of the impact of model size will be framed exclusively in descriptive terms.

Each model was tested on the full dataset using three evaluation methods: \textbf{string probabilities} (\textbf{SP}), \textbf{forced choice} (\textbf{FC}), and \textbf{yes/no} (\textbf{YN}). Each method involved evaluating the models based on various output token probabilities, whether they be for the sentences themselves (SP) or for potential metalinguistic judgments (FC/YN); these token probabilities were obtained using the \texttt{minicons} package (\citealt{minicons}).

\paragraph{String probabilities (SP)}

For the first method, whether a model got a minimal pair correct was determined by whether it assigned a higher log probability to the good sentence than to the bad sentence in a specific context $C$:
$$\log{P}(s_{\text{good}}\ \vert\ {C}) > \log{P}(s_{\text{bad}}\ \vert\ {C})$$

The context for a given item was determined as follows. All message histories started with a system message of ``You are a helpful assistant.'' If there was a background, it was encoded as a user message; if not, the user message was empty. If there was a prefix, then this was encoded as the beginning of an assistant message that the LM was to continue. For \textsc{only} and \textsc{adv\_qt} items, in addition to providing the background as a user message, the context included ``Therefore,\ '' as an assistant message prefix; this was done to encourage LM outputs that served as deductions based on the information in the background.

\paragraph{Forced choice (FC)}

The first prompting method we tested was a forced choice paradigm. The system prompt (``You are a helpful assistant.'') was followed by a user message with a brief introduction to the task, the relevant background or prefix, and both the good and bad follow-up sentences (Sentences A and B). The model was then tasked with responding with ``A'' or ``B'', based on which sentence was preferred. Two prompts were tested (FC1/FC2), with slight variations in both how the instructions were presented and whether the tested LM response was simply ``A''/``B'' or included an assistant message prefix of ``My answer is\ ''.

To control for a potential ordering bias (\citealt{PezeshkpourHruschka2024}), each item was run twice: once with the good sentence as Sentence A, and once as Sentence B. A model was considered to have gotten a given item correct if the summed log probabilities of the correct answers exceeded the summed log probabilities of the incorrect answers:
\begin{multline*}
\log{P}(\text{``A''}\ \vert\ C_{\text{A}}) + \log{P}(\text{``B''}\ \vert\ C_{\text{B}}) > {}\\
\log{P}(\text{``A''}\ \vert\ C_{\text{B}}) + \log{P}(\text{``B''}\ \vert\ C_{\text{A}})
\end{multline*}

% Note that in a certain sense this is a fairly generous metric, as it is possible for a model to respond ``A'' for every prompt---regardless of which sentence is the good one---and still receive an accuracy of 100\%, so long as the model's pro-``A'' bias decreases when the correct sentence is B.

\paragraph{Yes/No (YN)}

The other metalinguistic prompting method we used was a simple yes-or-no prompt, where the model was presented with one sentence at a time along with the background/prefix, and was asked to answer ``Yes'' or ``No'' depending on if the sentence was acceptable. We again tested this paradigm with two prompts (YN1/YN2), which varied similarly to FC's.

To decide the accuracy for a given item, we adopted a similar method as for FC: a model got an item correct if the summed log probabilities of accurate answers exceeded the summed log probabilities of inaccurate answers:
\begin{multline*}
\log{P}(\text{``Yes''}\ \vert\ C_{\text{good}}) + \log{P}(\text{``No''}\ \vert\ C_{\text{bad}}) > {}\\
\log{P}(\text{``Yes''}\ \vert\ C_{\text{bad}}) + \log{P}(\text{``No''}\ \vert\ C_{\text{good}})
\end{multline*}

The items in the dataset are designed so that the good and bad sentences are both acceptable in and of themselves, and it is only the background or prefix that determines which sentence is the better one. But there is a potential for bias in the dataset. Note that for a sentence like \emph{Mary greeted the vicar}, backgrounds could be devised that make the focusing of either \emph{the} or \emph{vicar} felicitous; however, the latter is more natural out of the blue. Thus, if a background is provided that encourages the focusing of \emph{vicar}, and a model gets that right, then there is a risk that the model is relying on a simpler syntactic heuristic dictating that nouns like \emph{vicar} are more likely to be focused than determiners like \emph{the}, rather than using the background information as desired (cf.\ \citealt{GururanganEtAl2018} for similar observations on NLI datasets). To test for such bias, a control version of each item was run with the background/prefix removed, for all models and evaluation methods. Whether or not there was a bias in the dataset could then be inferred based on whether models' performance significantly deviated from chance.

\subsection{Results}

\begin{figure*}[t]
  \begin{center}
    \noindent\includegraphics[width=.8\textwidth]{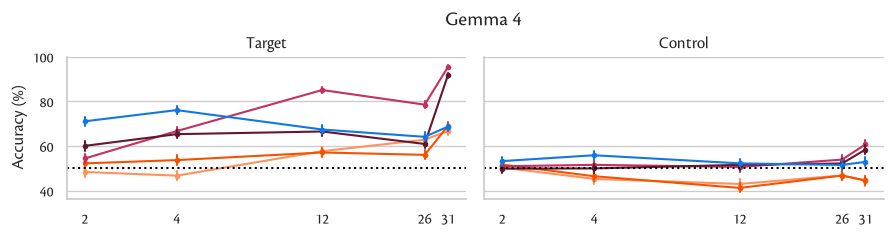}

    \noindent\includegraphics[width=.8\textwidth]{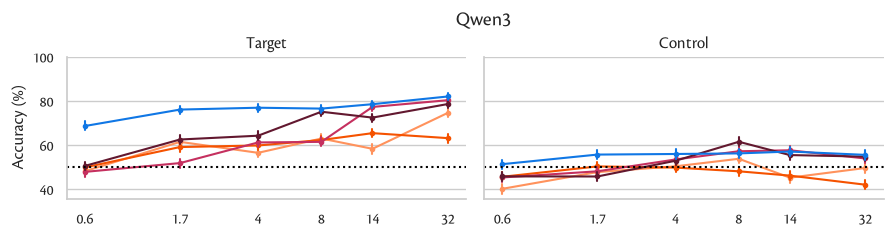}

    \noindent\includegraphics[width=.8\textwidth]{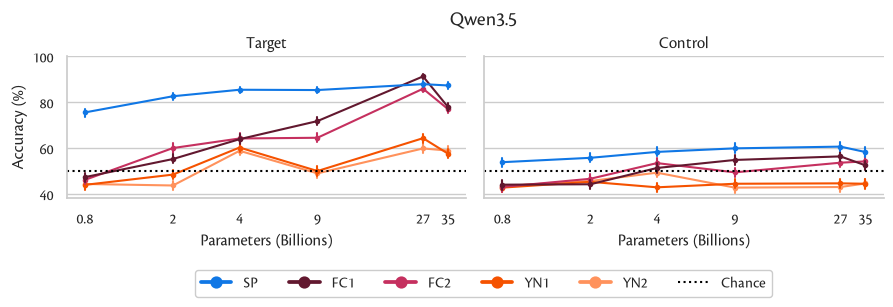}
  \end{center}
  \caption{Results of Experiment 1 (standard focus markers). Error bars are bootstrapped 95\% confidence intervals. X-axes are logarithmically spaced.\label{fig:exp-1-results}}
\end{figure*}

The results of Experiment 1 can be seen in Figure \ref{fig:exp-1-results}. Starting with the control items, larger models tended to perform slightly but significantly above chance, suggesting that there is in fact a non-trivial degree of bias in the dataset. Nonetheless, the results for the target items outpace the controls by a wide enough margin that we can safely conclude that positive target results are not attributable to mere bias, and the LMs' success is significantly dependent on the backgrounds and prefixes.

Moving on to the target items, many of the results for focus look similar to observations from prior work. To test for differences between methods, we fitted a mixed-effect logistic regression for each LM, predicting accuracy with a fixed effect of method (baseline = SP) and random effects for focus marker and item; statistical tests were performed using the \texttt{lme4} R package (\citealt{lme4}). SP performed above chance for all models ($p < .001$ across the board). For the smaller models, SP also significantly outperformed the prompting-based methods, in line with previous observations (\citealt{HuLevy2023,HuEtAl2024,KaufEtAl2024}). FC and YN generally improved as models got bigger, which fits neatly with prior results from \citet{HuFrank2024}.

Among prompting methods, FC generally outperformed YN, especially in larger models. This is unsurprising, since FC has the advantage of directly comparing the paired sentences, though we also cannot rule out the possibility that YN could improve with prompt engineering. FC outperforming YN also fits well with prior work on LMs (e.g., \citealt{PanicksseryEtAl2024}), as well as with research on humans, where forced choice results appear to be more powerful than single-item evaluations for both linguistic acceptability judgments (\citealt{SprouseEtAl2013}) and tasks in other cognitive domains (\citealt{GigerenzerRichter1990}).

However, a notable wrinkle in our results comes from those cases where prompting significantly outperformed string probabilities. This was observed for Qwen3.5 27B, where FC1 outperformed SP ($p < .001$), but the gap was much wider for Gemma 4 12B FC2, 26B-A4B FC2, and 31B FC1 and FC2 (all $p < .001$), each of which outperformed their respective SP scores by a considerable margin. In fact, among all model-method pairs, each of the top three performers used FC: Gemma 4 31B FC2 (95.3\%) and FC1 (91.7\%), and Qwen3.5 27B FC1 (91.3\%).

Note also that the two mixture-of-experts models, Gemma 4 26B-A4B and Qwen3.5 35B-A3B, saw a dip in performance compared to smaller non-MoE models in the same family. This suggests that MoE's reduction in active parameters has a negative impact on performance that is especially felt by prompting-based methods.

\begin{figure}[t]
  \begin{center}
    \noindent\includegraphics[width=.9\linewidth]{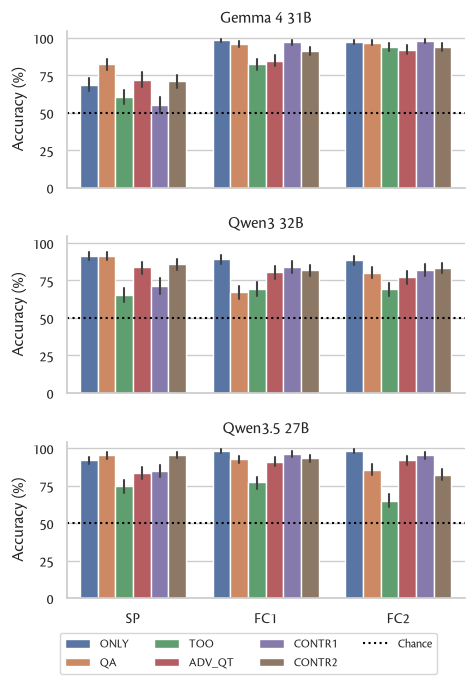}
  \end{center}
  \caption{SP/FC accuracy by type of focus-sensitivity for the top-performing model in each family. Error bars are bootstrapped 95\% confidence intervals.\label{fig:category-comp}}
\end{figure}

Figure \ref{fig:category-comp} zooms in on the best-performing model in each model family, showing SP and FC accuracy for each focus-sensitive environment. On a descriptive level, \textsc{only} items mostly saw high accuracy across models and methods, while \textsc{too} items showed lower accuracy. However, no definitive conclusions can be drawn from such observations, since manually created data could contain biases that make some items more difficult for independent reasons (\citealt{EttingerEtAl2018}). More importantly for our purposes, note that Gemma 4 31B FC1 and FC2 outperformed SP across all categories, suggesting that the advantage of FC was general and not confined to a subclass of environments. The picture is more mixed for Qwen3.5 27B FC1 (which also beat SP), partly due to SP's near-ceiling performance in multiple categories.

As was discussed above, the fact that SP outperforms prompting in smaller models is relatively unsurprising in light of previous work: SP is a direct reflection of the language modeling task, whereas metalinguistic judgments are a more demanding task that smaller models will tend to struggle with (\citealt{HuFrank2024}). The puzzle, then, is how FC can come to significantly \emph{outperform} SP, especially by the wide margins seen for Gemma 4. The answer, we propose, is that once FC is acquired, it is better equipped than SP to evaluate sentences that diverge substantially from normal model outputs. Recall that in the dataset focus marking is always out of the blue, as it is absent from the backgrounds and prefixes. It seems at least plausible that a result of the models' training would be a disinclination to produce, say, HTML <b>{\dots}</b> tags without solicitation. This could in turn muddy the waters when it comes to comparing the string probabilities of sentences that do contain such out-of-the-blue tags, which would exert downward pressure on SP accuracy.

FC, meanwhile, has two advantages over SP in this regard. First, the evaluated sentences (and their focus marking) are part of the input rather than the output, rendering irrelevant the low likelihood that the model would produce them as outputs. And second, FC is unlike SP and YN in that the model is presented with both sentences simultaneously, which introduces additional clarity in terms of both the criteria of evaluation and the bounds of comparison. The poor performance of YN---which enjoys the first advantage but not the second---suggests that the latter may be more impactful, though YN's poor performance could well be chalked up to independent factors.

The proposed picture, then, is as follows. First, the increased task demands of metalinguistic judgments entail that the success of prompting methods will be particularly dependent on model size. Hence, performance is poor for small models and improves for larger models. Second, when comparing between prompting methods, FC has a significant advantage over YN, since for FC the model sees both sentences simultaneously, while for YN the model must make each decision in isolation. And third, for those models that achieve strong prompting-based performance, FC has an advantage over SP when evaluating sentences that diverge substantially from normal model outputs.

This proposal makes a noteworthy prediction: if we modify our items so that they diverge \emph{even further} from normal model outputs, then any concomitant negative impact on accuracy should more strongly affect SP than FC. In other words, once FC judgments are successfully learned, they should be more robust to stylistic atypicality than SP. We test this prediction next.

\section{Experiment 2: \emph{Ad hoc} focus markers\label{sec:exp-2}}

In Experiment 2, we replace common focus markers like boldface and italics with randomly-generated \emph{ad hoc} focus markers. This serves two purposes. The first is to determine whether the models' knowledge of focus is specific to a closed class of canonical focus markers, or whether they have learned something more general. The second is to determine whether these more artificial focus markers lead to a greater decrease in accuracy for SP than for FC, as is predicted by our proposal.

\subsection{Methodology}

We generated ten \emph{ad hoc} focus markers. Each consisted of a randomly generated sequence of three to five characters sampled with replacement from the set $\{$``@'', ``\#'', ``\$'', ``\%'', ``;'', ``:''$\}$. This sequence was placed immediately before the focused constituent, with the reversed version of the sequence being placed immediately after. This is illustrated by the minimal pair in \eqref{ex:random-foc}:
\begin{exe}
  \ex   \begin{xlist}
          \ex   Lucy only \$:\#swam\#:\$ yesterday.
          \ex   Lucy only swam \$:\#yesterday\#:\$.
  \end{xlist}\label{ex:random-foc}
\end{exe}
% The intuition from a human perspective is that using randomly selected characters to bracket focused constituents adds ``weight'' to those constituents, thereby emulating the prosodic weight used to convey focus in spoken and signed language.
The total dataset therefore consisted of 10 $\times$ 282 $=$ 2,820 minimal pairs. Other than the change in focus marking, the experiment was run identically to Experiment 1.

\subsection{Results}

\begin{figure}[t]
  \begin{center}
    \noindent\includegraphics[width=.9\linewidth]{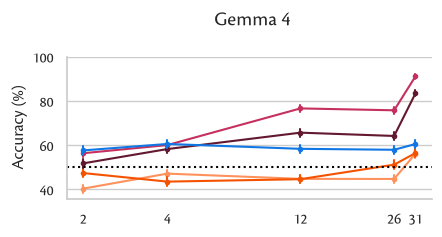}

    \noindent\includegraphics[width=.9\linewidth]{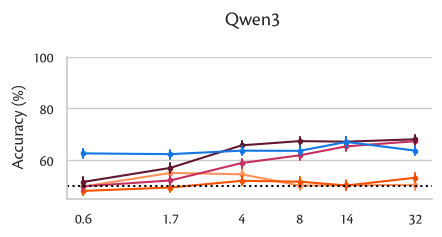}

    \noindent\includegraphics[width=.9\linewidth]{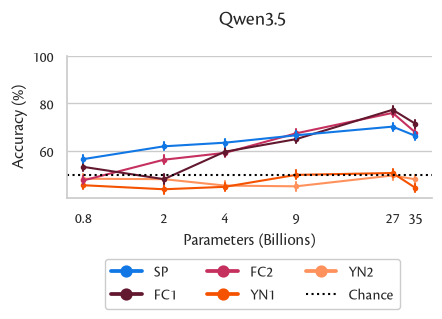}
  \end{center}
  \caption{Target results of Experiment 2 (\emph{ad hoc} focus-markers). Error bars are bootstrapped 95\% confidence intervals. X-axes are logarithmically spaced.\label{fig:exp-2-results}}
\end{figure}

The target item results for Experiment 2 can be seen in Figure \ref{fig:exp-2-results}. The control results, which we do not show for reasons of space, look similar to the control results from Experiment 1, but with lower peak accuracy. These continue to be outpaced by the target results, which again suggests that the backgrounds and prefixes play a crucial role. Put succinctly, the LMs are not just relying on a simple syntactic heuristic.

Turning to the target items, to compare methods we again fitted a mixed-effect logistic regression for each LM, predicting accuracy with a fixed effect of method (baseline = SP) and random effects of focus marker and item, using \texttt{lme4}. SP continued to perform significantly above chance for every tested model (all $p<.001$). This strongly suggests that what these models have learned about focus has a certain degree of generality, extending beyond canonical focus markers. The broad trend for FC also carried over from Experiment 1 to 2, with weak performance in smaller models that gradually improved to well beyond chance for larger, more capable models. The results for YN, which were the weakest in Experiment 1, more or less hovered around chance in Experiment 2.

\begin{figure}[t]
  \begin{center}
    \begin{tabular}{rrr}
      \includegraphics[scale=0.4]{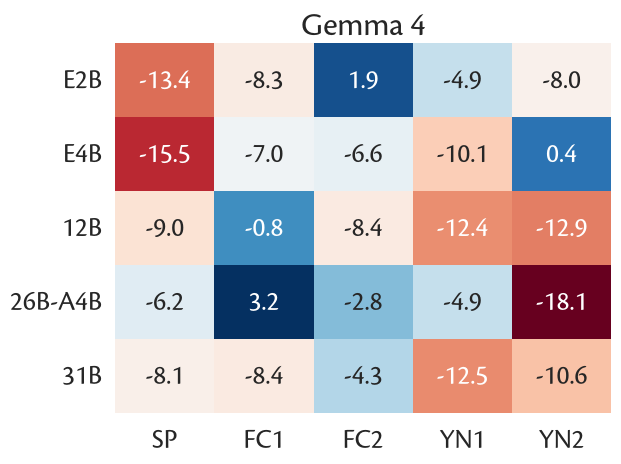}\\
      \includegraphics[scale=0.4]{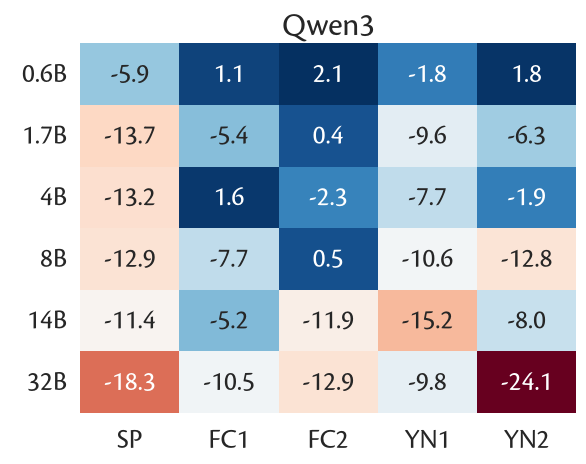}\\
      \includegraphics[scale=0.4]{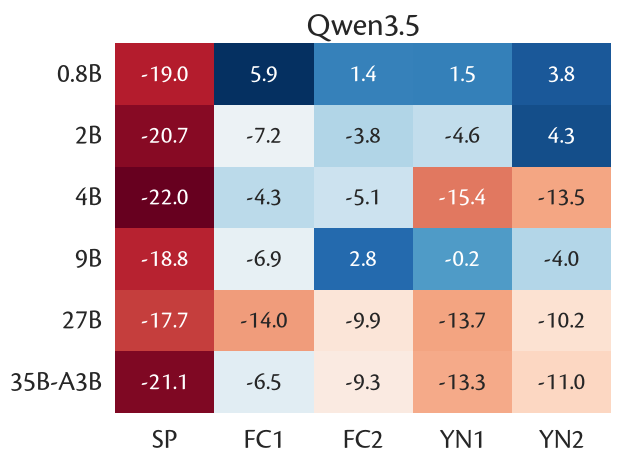}
    \end{tabular}
  \end{center}
  \caption{Percentage-point change in accuracy from Experiment 1 to Experiment 2 by model and method.\label{fig:accuracy-change}}
\end{figure}

Importantly, the results from Experiment 2 do show the behavior predicted in Section \ref{sec:exp-1}: FC performance degraded less than SP performance when moving from canonical to randomly generated focus markers. The change in percentage-point accuracy from Experiment 1 to Experiment 2 for each model and method can be seen in Figure \ref{fig:accuracy-change}; note that for nearly every model, both FC1 and FC2 had a less negative change than SP did.

This observation holds even when accounting for the fact that SP had ``farther to fall'', in the sense that SP often outperformed FC in Experiment 1. For example, when looking at the five pairs of model and prompting method that outperformed SP in Experiment 1, in four out of five cases FC still showed a smaller decrease in accuracy than SP did. Moreover, there are many more instances of FC significantly outperforming SP in Experiment 2 (twelve) than in Experiment 1 (five). Neither of these observations is to be expected if the sole reason for the greater decline in accuracy for SP was its greater accuracy in Experiment 1.

As a statistical test, for each experiment we took the mean accuracy across all focus markers for each combination of model, method, and item. We then fitted a mixed-effect linear regression model predicting \emph{ad hoc} focus marker accuracy, with fixed effects of standard focus marker accuracy and evaluation method (baseline = SP) and random effects for item and LM, using the \texttt{lmerTest} R package (\citealt{lmertest}). Since one of the fixed effects is standard focus marker accuracy, this test controls for the fact that some model-method combinations had farther to fall than others. There was a significant positive effect of standard focus marker accuracy ($p < .001$), as well as positive effects for both FC1 and FC2 compared to the baseline of SP (both $p<.001$), meaning that going from standard to \emph{ad hoc} focus markers had a less negative impact on FC than on SP. Finally, there was a significant negative effect for both YN1 and YN2 ($p < .001$ for both), though this is difficult to interpret due to the general poor performance of YN and the possibility that poor results could be attributed to faulty prompting.

\section{Discussion and conclusion\label{sec:discussion}}

Our findings support the proposal put forward above. SP outperforming prompting methods in smaller models (\citealt{HuLevy2023,HuEtAl2024,KaufEtAl2024}), with prompting performance improving for larger models (\citealt{HuFrank2024}), is predicted if metalinguistic judgments constitute a non-trivial auxiliary task. Meanwhile, FC can outperform SP in larger models---including by a wide margin for Gemma 4---at least in part because FC, once learned, is more robust in its evaluation of sentences that are stylistically unlikely outputs. Hence, when switching from standard to randomly generated focus markers, FC saw a smaller decrease in accuracy.

The strength of the conclusions we can draw from these results is somewhat limited by our focus on a singular empirical domain. A natural way to test the generality of these results would be to evaluate using other kinds of linguistic data that models might be similarly disinclined to generate, such as syntactically well-formed but semantically nonsensical sentences (cf.\ \citealt{GulordavaEtAl2018}); non-standard or -prestige dialects; socially unacceptable language use; and statements that are unambiguously false. Even beyond the particular hypotheses put forth in this paper, our results show that methodological comparisons can benefit from the investigation of linguistic phenomena beyond those covered by standard benchmarks like BLiMP (\citealt{WarstadtEtAl2020}).

Another promising path for future research would be to evaluate LMs' linguistic competence using prompting-based methods that are not expressly metalinguistic in nature. Tasks like code generation, for example, are likely better represented in LMs' training than metalinguistic judgments, so it is possible that the former will outperform the latter, particularly for smaller models. Not only might this lead to cleaner results, but it can also help to clarify which pros and cons of metalinguistic prompting are due specifically to the metalinguistic nature of the task.

Finally, it bears noting that the relative merits of prompting and string probabilities depend on more than just accuracy: they also depend on what these methods actually test, as it is not obvious that they test the same fundamental language abilities for a given model. A growing body of research is dedicated to discerning whether LMs are capable of \emph{introspection}, defined as a model's ability to predict or detect its own outputs via prompting. On the one hand, positive evidence exists suggesting that some models can non-trivially recognize their own outputs (\citealt{PanicksseryEtAl2024}), describe implicit behavioral patterns acquired via finetuning (\citealt{BetleyEtAl2025}), and learn with disproportionate success to predict their own hypothetical outputs (\citealt{BinderEtAl2025}). On the other hand, \citet{SongEtAl2025}, who specifically test for introspection in metalinguistic judgments, fail to find it, though it is worth emphasizing that they base their conclusions of non-introspection on null results.

But suppose that \citet{SongEtAl2025} are right and LMs do not introspect when performing metalinguistic judgments, even when they judge with high accuracy. That is, the language knowledge LMs use to form judgments is distinct from the knowledge they use to generate tokens. One possible interpretation of such a result is that the explicit knowledge used by LMs to make judgments is essentially factual in nature, akin to knowing a bread recipe or a chess opening, whereas the implicit knowledge assessed via string probabilities is a more meaningful indicator of an LM's linguistic competence. If this is true, then the accuracy of prompting is irrelevant, since it is simply not testing the relevant kind of language ability.

But there is an alternate interpretation that does not relegate prompting methods to such second-class status. The core language modeling task could be (somewhat simplistically) broken down into the subtasks of (\textsc{i}) interpreting the model's input and (\textsc{ii}) generating a corresponding output. Importantly, there is no inherent reason to believe that the knowledge an LM uses to interpret a sentence $S$ as an input is identical to the knowledge it uses to produce $S$ as an output (\citealt{WestEtAl2024}).

From this perspective, there are really two ways in which metalinguistic prompting and string probabilities differ from each other: explicit vs.\ implicit and input vs.\ output. Explicit judgments are still difficult, hence the poor performance in smaller models. But for models that can overcome that, the judgments they provide are informed by a facet of their language knowledge that is distinct from what is used for generation---hence, not introspective in the technical sense---but no less important in terms of the fundamental language modeling task. Or, indulging in some anthropomorphizing terminology, prompting tests comprehension, while probabilities test production. If this is the case, then prompting and probabilities are not so much competitors as they are complements, and both are needed for a complete understanding of LMs' language competence.

All this being said, it remains to be seen whether introspection is in fact involved in LMs' metalinguistic judgments, and if so, where and to what extent. But it is important to bear in mind as work on introspection progresses that the potential value of prompting-based linguistic assessment does not hinge entirely on whether LMs' judgments are introspective in the particular sense defined above.

\iftaclpubformat
\section*{Acknowledgments}

This work is supported by the German Research Foundation, project \#579297085 (\emph{Propositional Attitudes in Large Language Models}), funded as part of the LaSTing Priority Program. Many thanks to members of the LaSTing community and to colleagues at ZAS for fruitful discussion.
\fi

\bibliography{references}
\bibliographystyle{acl_natbib}

\onecolumn

\appendix

\end{document}